\documentclass[10pt,journal,compsoc]{IEEEtran}
\ifCLASSOPTIONcompsoc
  \usepackage[nocompress]{cite}
\else
  \usepackage{cite}
\fi

\usepackage{ragged2e}
\usepackage{epsfig}

\usepackage{algorithmic}
\usepackage{url}
\usepackage{enumerate}
\usepackage{bm}
\usepackage{multirow}
\renewcommand\figurename{Figure}
\usepackage{graphicx}
\usepackage[ruled]{algorithm2e}
\usepackage{amsmath}
\usepackage{CJK}

\SetAlFnt{\small}
\SetAlCapFnt{\small}
\SetAlCapNameFnt{\small}
\SetAlCapHSkip{0pt}
\IncMargin{-\parindent}
\DeclareMathSizes{14}{14}{9.8}{7}

\ifCLASSINFOpdf
\else

\fi
\begin{document}

%
\title{$\rho$-hot Lexicon Embedding-based Two-level LSTM for Sentiment Analysis}

\author{Ou~Wu,~Tao~Yang,~Mengyang~Li,~and~Ming~Li
\IEEEcompsocitemizethanks{\IEEEcompsocthanksitem Ou Wu is with Center for Applied Mathematics, Tianjin University, E-mail: wuou@tju.edu.cn
\IEEEcompsocthanksitem Tao Yang, Mengyang Li, and Ming Li are with Civil Aviation University of China, E-mail: \{yangtao087, liming2960, limengyang99\}@gmail.com.}
\thanks{All authors contributed equally. Ou Wu initiated the research and proposed the lexicon embedding part. Tao Yang and Mengyang Li proposed the labeling strategy and the two-level LSTM. Ming Li implemented the whole approach and performed the experimental analysis. Ou Wu wrote the paper.	All the experimental data sets and codes are publicly available at Github: https://github.com/Tju-AI/two-stage-labeling-for-the-sentiment-orientations}}

\markboth{Journal of \LaTeX\ Class Files,~Vol.~14, No.~8, August~2015}%
{Shell \MakeLowercase{\textit{et al.}}: Bare Demo of IEEEtran.cls for Computer Society Journals}

\IEEEtitleabstractindextext{%
\begin{abstract}
Sentiment analysis is a key component in various text mining applications. Numerous sentiment classification techniques, including conventional and deep learning-based methods, have been proposed in the literature. In most existing methods, a high-quality training set is assumed to be given. Nevertheless, constructing a high-quality training set that consists of highly accurate labels is challenging in real applications. This difficulty stems from the fact that text samples usually contain complex sentiment representations, and their annotation is subjective. We address this challenge in this study by leveraging a new labeling strategy and utilizing a two-level long short-term memory network to construct a sentiment classifier. Lexical cues are useful for sentiment analysis, and they have been utilized in conventional studies. For example, polar and privative words play important roles in sentiment analysis. A new encoding strategy, that is, $\rho$-hot encoding, is proposed to alleviate the drawbacks of one-hot encoding and thus effectively incorporate useful lexical cues. We compile three Chinese data sets on the basis of our label strategy and proposed methodology. Experiments on the three data sets demonstrate that the proposed method outperforms state-of-the-art algorithms.
\end{abstract}

}

\maketitle

\IEEEdisplaynontitleabstractindextext

%
\IEEEpeerreviewmaketitle

\IEEEraisesectionheading{\section{Introduction}\label{sec:introduction}}

%
%
%
%
\IEEEPARstart{T}{ext} is important in many artificial intelligence applications. Among various text mining techniques, sentiment analysis is a key component in applications such as public opinion monitoring and comparative analysis. Sentiment analysis can be divided into three problems according to input texts, namely, sentence, paragraph, and document levels. This study focuses on sentence and paragraph levels.

Text sentiment analysis is usually considered a text classification problem. Almost all existing text classification techniques are applied to text sentiment analysis \cite{LiuBook2015}. Typical techniques include bag-of-words (BOW)-based \cite{Alexander17}, deep learning-based \cite{Kim1746}, and lexicon-based (or rule-based) methods \cite{Taboada267}.

Although many achievements have been made and sentiment analysis has been successfully used in various commercial applications, its accuracy can be further improved. The construction of a high-accuracy sentiment classification model usually entails the challenging compilation of training sets with numerous samples and sufficiently accurate labels. The reason behind this difficulty is two-fold. First, sentiment is somewhat subjective, and a sample may receive different labels from different users. Second, some texts contain complex sentiment representations, and a single label is difficult to provide. We conduct a statistical analysis of public Chinese sentiment text sets in GitHub. The results show that the average label error is larger than 10\%. This error value reflects the degree of difficulty of sentiment labeling.

Privative and interrogative sentences are difficult to classify when deep learning-based methods are applied. Although lexicon-based methods can deal with particular types of privative sentences, their generalization capability is poor.

We address the above issues with a new methodology. First, we introduce a two-stage labeling strategy for sentiment texts. In the first stage, annotators are invited to label a large number of short texts with relatively pure sentiment orientations. Each sample is labeled by only one annotator. In the second stage, a relatively small number of text samples with mixed sentiment orientations are annotated, and each sample is labeled by multiple annotators. Second, we propose a two-level long short-term memory (LSTM) \cite{Hochreiter1735} network to achieve two-level feature representation and classify the sentiment orientations of a text sample to utilize two labeled data sets. Lastly, in the proposed two-level LSTM network, lexicon embedding is leveraged to incorporate linguistic features used in lexicon-based methods.

Three Chinese sentiment data sets are compiled to investigate the performance of the proposed methodology. The experimental results demonstrate the effectiveness of the proposed methods. Our work is new in the following aspects.

\begin{itemize}
  \item A highly effective labeling strategy is adopted. Labeling a high-quality training set is difficult in sentiment analysis. In our labeling strategy, samples are divided into ones with relatively pure sentiment orientations and ones with relatively complex sentiment orientations. This procedure is easily performed in practice.
  \item A two-level LSTM network is proposed. Our labeling procedure yields two training sets with different sentiment levels; therefore, we propose a two-level LSTM network that can effectively utilize the two data sets.
  \item Lexicon embeddings are introduced. We introduce a new embedding method to incorporate useful cues usually used in lexicon-based methods and to ultimately utilize lexical cues.
  \item A new encoding strategy is introduced. One-hot encoding is the most widely used encoding strategy for categorical data. An effective encoding strategy, $\rho$-hot encoding, is proposed in this work to address the limitations of one-hot encoding.
\end{itemize}

The rest of this paper is organized as follows. Section 2 briefly reviews related work. Section 3 describes our methodology. Section 4 reports the experimental results, and Section 5 concludes the study.

\section{RELATED WORK}

\subsection{Text Sentiment Analysis}
Sentiment analysis aims to predict the sentiment polarity of an input text sample. Sentiment polarity can be divided into negative, neutral, and positive in many applications.

Existing sentiment classification methods can be roughly divided into two categories, namely, lexicon-based and machine learning-based methods \cite{ZhaoGCHCWW18}. Lexicon-based methods \cite{Hu168} construct polar and privative word dictionaries. A set of rules for polar and privative words is compiled to judge the sentiment orientation of a text document. This method cannot effectively predict implicit orientations. Machine learning-based methods \cite{Mullen412} utilize a standard binary or multi-category classification approach. Different feature extraction algorithms, including BOW \cite{Paltoglou1386} and part of speech (POS) \cite{Mullen412}, are used. Word embedding and deep neural networks have recently been applied to sentiment analysis, and promising results have been obtained \cite{Glorot513}\cite{Tang3298}.

\begin{figure}[htbp]
\begin{center}
  \hspace{0in}\includegraphics[width= 0.49\textwidth, height = 150pt]{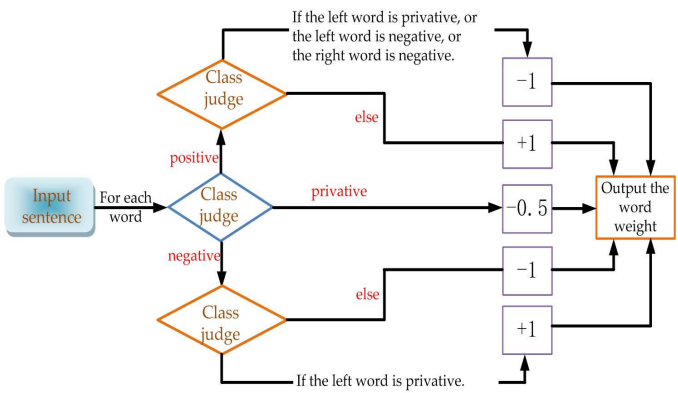}\\
  \vspace{-0.05in}\caption{A lexicon-based approach for sentiment classification.}\vspace{-0.15in}
\end{center}
\end{figure}


\subsection{Lexion-based Sentiment Classification}
Lexicon-based methods are actually in implemented in an unsupervised manner. They infer the sentiment categories of input texts on the basis of polar and privative words. The primary advantage of these methods is that they do not require labeled training data. The key of lexicon-based methods is the lexical resource construction, which maps words into a category (positive, negative, neutral, or privative). Senti-WordNet \cite{Baccianella2200} is a lexical resource for English text sentiment classification. For Chinese texts, Senti-HowNet is usually used.

Fig. 1 characterizes a typical lexicon-based sentiment classification approach. The approach iteratively checks each word in an input sentence from left to right. The weight score of each word is calculated according to the procedure shown in Fig. 1. The final sentiment score is the average score of the words with weight scores. The scores of positive, neutral, and negative sentiments are denoted as ``+1",``0", and ``-1", respectively. According to the lexicon-based algorithm shown in Fig. 1, the sentiment score of ``it is not bad" is 0.25, and the sentiment score of ``it is good" is 1. However, the score of ``it is not so bad" is -0.75, and this score is definitely wrong. Therefore, machine learning (including feature learning) methodologies have become mainstream in sentiment analysis.

\subsection{Deep Learning-based Sentiment Classification}
Deep learning (including word embedding \cite{Lai5}) has been applied to almost all text-related applications, such as translation \cite{Cho1724}, quality assurance \cite{Yang645}, recommendation \cite{Dai1507}, and categorization \cite{Tang1422}. Popular deep neural networks are divided into convolutional neural networks (CNNs) \cite{Kalchbrenner655} and recurrent neural network (RNNs) \cite{Socher129}\cite{Zhang3087}. Both are utilized in sentiment classification \cite{Santos69}. Kim investigated the use of CNN in sentence sentiment classification and achieved promising results \cite{Kim1746}. LSTM \cite{Gers2451}, a classical type of RNN, is the most popular network used for sentiment classification. A binary-directional LSTM \cite{Zhou247} with an attention mechanism is demonstrated to be effective in sentiment analysis.

Deep learning-based methods rarely utilize the useful resources adopted in lexicon-based methods. Qiao et al. \cite{Qian1611} incorporated lexicon-based cues into the training of an LSTM-based model. Their proposed method relies on a new loss function that considers the relationships between polar or certain types of words (e.g., privative) and those words next to them in input texts. Our study also combines lexical cues into LSTM. Nevertheless, unlike Qiao et al.'s study that implicitly used lexical cues, the present work explicitly uses lexical cues in the LSTM network.
Shin et al. \cite{Shin149} combined the lexicon embeddings of polar words with word embeddings for sentiment classification. The difference between our approach an the method proposed by Shin et al. the is discussed in Section 3.3.5.

Numerous studies on aspect-level sentiment analysis exist \cite{Schouten813}. This problem is different from the sentiment classification investigated in this study.

 \begin{figure*}[t]
\begin{center}
  \hspace{0in}\includegraphics[width= 0.786\textwidth]{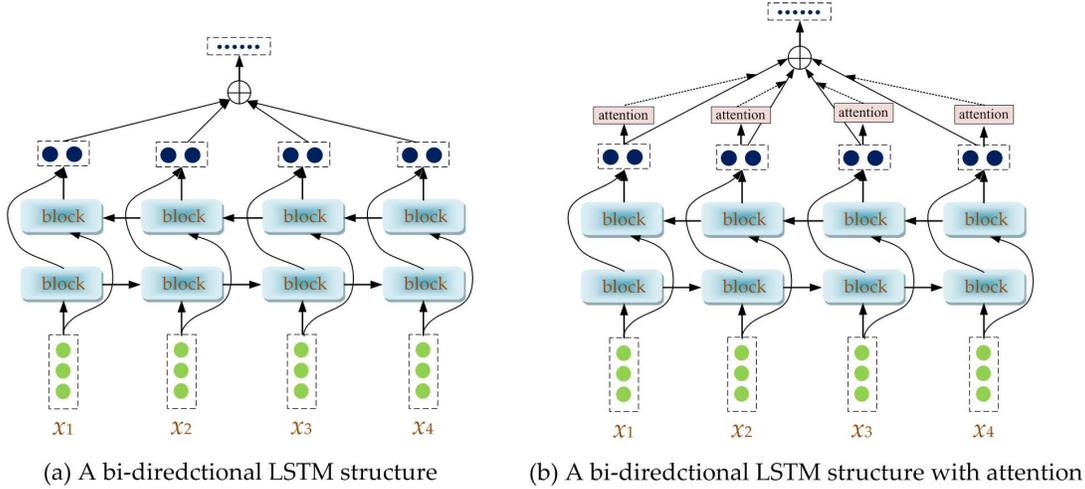}\\
  \vspace{-0.05in}\caption{Two representative LSTM structures for text classification: a bi-directional LSTM (left), and a bi-directional LSTM with attention (right).}\vspace{-0.15in}
\end{center}
\end{figure*}

\section{METHODOLOGY}
This section first introduces our two-stage labeling procedure. A two-level LSTM is then proposed. Lexicon embedding is finally leveraged to incorporate lexical cues.
\subsection{Two-stage Labeling}
As stated earlier, sentiment is subjective, and texts usually contain mixed sentiment orientations. Therefore, texts’ sentiment orientations are difficult to label. In our study, three sentiment labels, namely, positive, neutral, and negative, are used. The following sentences are taken as examples.
\begin{description}
  \item[\textbf{S1}:] The service is poor. The taste is good, but the rest is not so bad.
  \item[\textbf{S2}:] The quality of the phone is good, but the appearance is just so-so.
\end{description}

In user annotation, the labels of these two sentences depend on users. If a user is concerned about service, then the label of S1 may be ``negative". By contrast, for another user who does not care about service, the label may be ``positive". Similarly, a user may label S2 as ``positive" if he cares about quality. Another user may label it as ``negative" if the conjunction ``but" attracts the user’s attention more. Another user may label it as ``neutral" if they are concerned about quality and appearance.

The underlying reason is that sentiment is more subjective than semantics. In related research on subjective categorization, such as visual aesthetics, each sample is usually repeatedly annotated by multiple annotators, and the average label is taken as the final label of the sample. This labeling strategy can also be applied to text sentiment annotation. However, we argue that this strategy is unsuitable for a (relatively) large number of samples. The reason lies in the following two aspects.

\begin{itemize}
  \item Multiple annotators for a large number of data sets require a large budget.
  \item In our practice, annotators claim that their judgment criteria on sentiment become fused on texts with mixed sentiment orientations (e.g., S1 and S2) over time during labeling, and they become bored accordingly.
\end{itemize}

A two-stage labeling strategy is adopted in this study. In the first stage, each sentence/paragraph is divided into several clauses according to punctuation. The sentiment of each partitioned clause is relatively easy to annotate; therefore, each clause is labeled by only one user. In the second stage, a relatively small-sized sentence/paragraph set is labeled, and each sentence is labeled by all involved annotators. We still take the two sentences, S1 and S2, as examples. S1 and S2 are split into clauses, as shown below.
\begin{itemize}
  \item S1:
  \begin{itemize}
    \item S1.1: The service is poor
    \item S1.2: The taste is good
    \item S1.3: but the rest is not so bad.
  \end{itemize}
  \item S2:
  \begin{itemize}
    \item S2.1: The quality of the phone is good
    \item S2.2: but the appearance is just so-so.
  \end{itemize}
\end{itemize}
Each of the above clauses is labeled by only one annotator. The annotation in the first stage is easy to perform; thus, the number of clauses can be larger than the number of sentences used in the second labeling stage.


\subsection{Two-level LSTM}
Given two training data sets (denoted by T1 and T2), a new learning model should be utilized. LSTM\footnote{CNN is another widely used text classification model. Our idea can also be applied to CNN.} is a widely used deep neural network in deep learning-based text classification.

LSTM is a typical RNN model for short-term memory, which can last for a long period of time. An LSTM is applicable to classify, process, and predict time series information with given time lags of unknown size. A common LSTM block is composed of a cell, an input gate, an output gate, and a forget gate. The forward computation of an LSTM block at time $t$ or position $t$ is as follows \cite{Gers2451}:
\begin{equation}
\begin{array}{l}
{i_t} = \sigma ({W_i}[{c_{t - 1}},{h_{t - 1}},{x_t}] + {b_i})\\
{f_t} = \sigma ({W_f}[{c_{t - 1}},{h_{t - 1}},{x_t}] + {b_f})\\
{o_t} = \sigma ({W_o}[{c_t},{h_{t - 1}},{x_t}] + {b_o})\\
{d_t} = \sigma ({W_d}[{c_{t - 1}},{h_{t - 1}},{x_t}] + {b_d})\\
{c_t} = {i_t} \otimes {d_t} + {f_t} \otimes {d_{t - 1}}\\
{h_t} = {o_t} \otimes \tanh ({c_t})
\end{array}
\end{equation}
where $x_t$ is the input vector at time $t$ (or position $t$); $i_t$ and $d_t$ are the input vectors of the input unit and input gate, respectively; $o_t$ and $h_t$ are the output and hidden vectors at time $t$, respectively; $f_t$ is the output of the forget gate at time $t$; $c_t$ is the internal state of the memory cell in an LSTM block at time $t$; and $\sigma$ is the sigmoid active function.

When LSTM is used to classify an input sentence, the hidden vectors of each input vector are summed to form a dense vector that can be considered the feature representation of the input sentence, i.e.,
\begin{equation}
{v_t} = \sum\limits_t {{h_t}}
\end{equation}

 \begin{figure*}[ht]
\begin{center}
  \hspace{0in}\includegraphics[width= 0.88\textwidth]{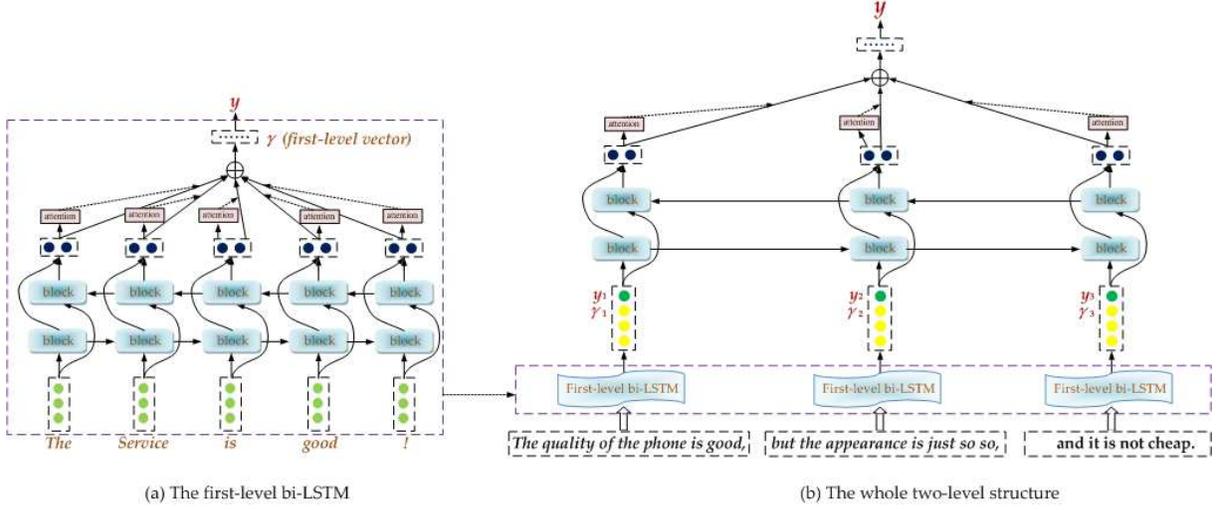}\\
  \vspace{-0.05in}\caption{The proposed two-level LSTM network.}\vspace{-0.15in}
\end{center}
\end{figure*}

In many applications, a bi-directional LSTM (bi-LSTM) structure is usually used, as shown in Fig. 2(a). In bi-LSTM, forward and backward information are considered for information at time $t$; hence, the context is modeled. Bi-LSTM is thus significantly reasonable for text processing tasks. In our two-level LSTM, bi-LSTM is used in each level.

The output hidden state at time $t$ of a bi-LSTM block can be described as follows:

\begin{equation}
\begin{array}{l}
\overrightarrow{h}_t   = \overrightarrow{o}_t \otimes \tanh (\overrightarrow{c}_t)\\
\overleftarrow{h}_t = {\overleftarrow{o}_t} \otimes \tanh (\overleftarrow{c}_t)\\
{h_t} = [\overrightarrow{h}_t,\overleftarrow{h}_t]
\end{array}
\end{equation}
where $\overrightarrow{h}_t$, $\overrightarrow{o}_t$, and $\overrightarrow{c}_t$ are the corresponding vectors at time $t$ in the forward LSTM block; and $\overleftarrow{h}_t$, $\overleftarrow{o}_t$, and $\overleftarrow{c}_t$ are the corresponding vectors at time $t$ in the backward LSTM block. $H = \{h_t, \cdots, h_T\}$. When attention is used, the dense feature vector $\gamma$ of an input sentence is calculated as follows:
\begin{equation}
\begin{array}{l}
\alpha {\rm{ = }}{\mathop{\rm softmax}\nolimits} {\rm{(}}{{\rm{\bf{w}}}^T}H{\rm{)}}\\
{\rm{\gamma  =  }}H{\alpha ^T}
\end{array}
\end{equation}
where $\alpha$ is the vector that consists of attention weights. The bi-LSTM with attention is shown in Fig. 2(b).

Our proposed network consists of two levels of LSTM network. In the first level, a bi-LSTM is used and learned on the basis of the first training set T1. This level is a conventional sentiment classification process. The input of this level is a clause, and the input $x_t$ is the embedding of the basic unit of the input texts\footnote{In English texts, the basic unit is usually a word; in Chinese texts, the basic unit is a Chinese word or character.}. The network is shown in Fig. 3(a).

In the second level, a bi-LSTM is also used and learned on the basis of the second training set T2. The input of this level is a sentence or a paragraph. The input $x_t$ consists of two parts\footnote{The third part is lexicon embedding which will be introduced in Section 3.3.4.}. The first part is the feature vector of the $t$-th clause. The feature vector is generated by the first-level network. In other words, the dense feature shown in Fig. 3(a) ($\gamma$) is used. The second part is the sentiment score (not predicted label) output by the first-level network. The sentence S1 (The service is poor. The taste is good, but the rest is not so bad.) used in Subsection 3.1 is taken as an illustrative example. S1 contains three clauses. Therefore, the input vector of S1 can be represented by
\[S1:{\rm{ }}\{{\eta_1},{\eta_2},{\eta_3}\}\]
where
\begin{equation}
\begin{array}{l}
{\eta _1} = \{ y_1^{(1)},\gamma _1^{(1)}\} \\
{\eta _2} = \{ y_2^{(1)},\gamma _2^{(1)}\} \\
{\eta _3} = \{ y_3^{(1)},\gamma _3^{(1)}\}
\end{array}
\end{equation}
where $y_i^{(1)}$ is the output score of the $i$th clause by the first-level LSTM and $\gamma_i^{(1)}$ is the feature representation of the $i$th clause by the first LSTM. The network of the whole two-level network is shown in Fig. 3(b).

\subsection{Lexical Embedding}
The proposed lexicon embedding is based on $\rho$-hot encoding. Therefore, $\rho$-hot encoding is first described.

\subsubsection{$\rho$-hot encoding}
For categorical data, one-hot encoding is the most widely used encoding strategy when different categories are independent\footnote{When different categories are correlated, sophisticated encoding strategies can be utilized. For example, one-hot is a traditional encoding for words. Many word-embedding methods are proposed with consideration of the relation among words.}. For example, if one-hot encoding is used to represent three categories, namely, positive, neutral, and negative, the encoding vectors for the three categories are $[1, 0, 0]^T$, $[0, 1, 0]^T$, and $[0, 0, 1]^T$, respectively.

In this work, many lexical cues are categorical data, and different categories are independent. These lexical cues can directly be represented by one-hot encoding. The encoded vectors for lexical cues are then concatenated with other vectors, such as character/word embedding. However, one-hot encoding presents two main limitations when the encoded vector is concatenated with other vectors.
\begin{itemize}
  \item The value difference between the elements of one-hot encoded vectors and those of other encoded vectors (e.g., word embedding vectors) may be large. Fig. 4 shows the histogram of the values of the elements of the word embedding vectors. The magnitude of most elements are smaller than 1.
  \item The lengths of one-hot encoded vectors are usually shorter than those of other encoded vectors. Consequently, the proportion of one-hot encoded part is small in the concatenated vectors.
\end{itemize}

The above two limitations affect the final sentiment analysis performance. To this end, we propose a new encoding strategy.
\begin{equation}
\rho  - {\bf{hot \quad}}{\rm{ }}{\bf{encoding}}:{\nu _\rho }{\rm{ = }}\rho  \cdot {\nu _{\bf{1}}} \otimes {{\bf{1}}_{n\times 1}}
\end{equation}
where $\nu_\rho$ is the $\rho$-hot encoded vector, $\rho$ is the proportion parameter, $\nu_1$ is the one-hot encoded vector, and $\textbf{1}_{n\times1}$ is an $n$-dimensional vector. If $\rho$ and $n$ are equal to 1, then $\rho$-hot encoding is reduced to one-hot encoding. The parameter $n$ is applied to increase the length of the final encoded vector.

\begin{figure}[t]
\begin{center}
  \hspace{0in}\includegraphics[width= 0.426\textwidth]{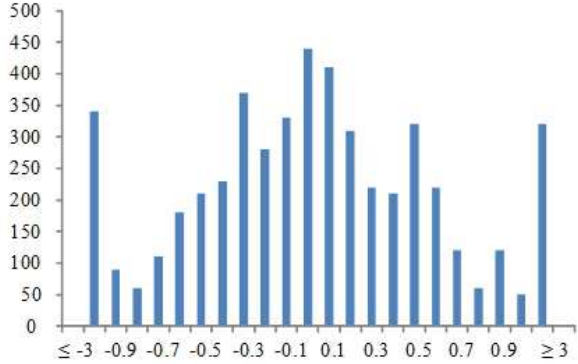}\\
  \vspace{-0.05in}\caption{The histogram of the values in word embedding vectors. Most values are smaller than 1.}\vspace{-0.15in}
\end{center}
\end{figure}

\subsubsection{Embedding for key lexical words}
Most lexicon-based sentiment methods rely on four types of words, namely, positive, negative, neutral, and privative. These words are useful cues for predicting the sentiment labels of input texts. The incorporation of these words should also be useful. A previous study has shown that a typical document comprises approximately 8\% of such sentences \cite{Narayanan180}. Sentiments expressed in a conditional sentence can be difficult to determine due to the semantic condition. The sentiment polarities of interrogative sentences are also difficult to classify according to our empirical study.

Five types of words, namely, positive (Pos), negative (Neg), privative (Pri), suppositive (Sup), and interrogative (Int), are represented by the proposed encoding method. The rest words, which do not belong to any of the above five types, are named ``others (Oth)" instead of ``neutral" because some words, such as ``the", are unrelated to ``sentiment". The value of $n$ in Eq. (6) is set as 10. The encoded vectors are as follows.
\[\begin{array}{l}
{\rm{ Pos}}:{[{{\bm{\rho }}_{1 \times 10}},{{\bf{0}}_{1 \times 10}},{{\bf{0}}_{1 \times 10}},{{\bf{0}}_{1 \times 10}},{{\bf{0}}_{1 \times 10}},{{\bf{0}}_{1 \times 10}}]^T}\\
{\rm{Neg}}:{[{{\bf{0}}_{1 \times 10}},{{\bm{\rho }}_{1 \times 10}},{{\bf{0}}_{1 \times 10}},{{\bf{0}}_{1 \times 10}},{{\bf{0}}_{1 \times 10}},{{\bf{0}}_{1 \times 10}}]^T}\\
{\rm{  Pri}}:{[{{\bf{0}}_{1 \times 10}},{{\bf{0}}_{1 \times 10}},{{\bm{\rho }}_{1 \times 10}},{{\bf{0}}_{1 \times 10}},{{\bf{0}}_{1 \times 10}},{{\bf{0}}_{1 \times 10}}]^T}\\
{\rm{ Sup}}:{[{{\bf{0}}_{1 \times 10}},{{\bf{0}}_{1 \times 10}},{{\bf{0}}_{1 \times 10}},{{\bm{\rho }}_{1 \times 10}},{{\bf{0}}_{1 \times 10}},{{\bf{0}}_{1 \times 10}}]^T}\\
{\rm{   Int}}:{[{{\bf{0}}_{1 \times 10}},{{\bf{0}}_{1 \times 10}},{{\bf{0}}_{1 \times 10}},{{\bf{0}}_{1 \times 10}},{{\bm{\rho }}_{1 \times 10}},{{\bf{0}}_{1 \times 10}}]^T}\\
{\rm{ Oth}}:{[{{\bf{0}}_{1 \times 10}},{{\bf{0}}_{1 \times 10}},{{\bf{0}}_{1 \times 10}},{{\bf{0}}_{1 \times 10}},{{\bf{0}}_{1 \times 10}},{{\bm{\rho }}_{1 \times 10}}]^T}
\end{array}\]

In the proposed $\rho$-hot embedding, the parameter $\rho$ can be learned during training. The representation of the third clause (``but the rest is not so bad") of S1 in Subsection 3.1 is taken as an illustrative example. The new embedding of each word in this clause is as follows.

\begin{small}
\begin{equation}
\left[ {\begin{array}{*{20}{c}}
{\phi(but)}&{\phi(the)}&{\phi(rest)}&{\phi(is)}&{\phi(not)}&{\phi(so)}&{\phi(bad)}\\
{{{\bf{0}}^{10{\rm{x}}1}}}&{{{\bf{0}}^{10{\rm{x}}1}}}&{{{\bf{0}}^{10{\rm{x}}1}}}&{{{\bf{0}}^{10{\rm{x}}1}}}&{{{\bf{0}}^{10{\rm{x}}1}}}&{{{\bf{0}}^{10{\rm{x}}1}}}&{{{\bf{0}}^{10{\rm{x}}1}}}\\
{{{\bf{0}}^{10{\rm{x}}1}}}&{{{\bf{0}}^{10{\rm{x}}1}}}&{{{\bf{0}}^{10{\rm{x}}1}}}&{{{\bf{0}}^{10{\rm{x}}1}}}&{{{\bf{0}}^{10{\rm{x}}1}}}&{{{\bf{0}}^{10{\rm{x}}1}}}&{{{\bm{\rho }}^{10{\rm{x}}1}}}\\
{{{\bf{0}}^{10{\rm{x}}1}}}&{{{\bf{0}}^{10{\rm{x}}1}}}&{{{\bf{0}}^{10{\rm{x}}1}}}&{{{\bf{0}}^{10{\rm{x}}1}}}&{{{\bm{\rho }}^{10{\rm{x}}1}}}&{{{\bf{0}}^{10{\rm{x}}1}}}&{{{\bf{0}}^{10{\rm{x}}1}}}\\
\begin{array}{l}
{{\bf{0}}^{10{\rm{x}}1}}\\
{{\bf{0}}^{10{\rm{x}}1}}\\
{{\bm{\rho }}^{10{\rm{x}}1}}
\end{array}&\begin{array}{l}
{{\bf{0}}^{10{\rm{x}}1}}\\
{{\bf{0}}^{10{\rm{x}}1}}\\
{{\bm{\rho }}^{10{\rm{x}}1}}
\end{array}&\begin{array}{l}
{{\bf{0}}^{10{\rm{x}}1}}\\
{{\bf{0}}^{10{\rm{x}}1}}\\
{{\bm{\rho }}^{10{\rm{x}}1}}
\end{array}&\begin{array}{l}
{{\bf{0}}^{10{\rm{x}}1}}\\
{{\bf{0}}^{10{\rm{x}}1}}\\
{{\bm{\rho }}^{10{\rm{x}}1}}
\end{array}&\begin{array}{l}
{{\bf{0}}^{10{\rm{x}}1}}\\
{{\bf{0}}^{10{\rm{x}}1}}\\
{{\bf{0}}^{10{\rm{x}}1}}
\end{array}&\begin{array}{l}
{{\bf{0}}^{10{\rm{x}}1}}\\
{{\bf{0}}^{10{\rm{x}}1}}\\
{{\bm{\rho }}^{10{\rm{x}}1}}
\end{array}&\begin{array}{l}
{{\bf{0}}^{10{\rm{x}}1}}\\
{{\bf{0}}^{10{\rm{x}}1}}\\
{{\bf{0}}^{10{\rm{x}}1}}
\end{array}
\end{array}} \right]
\end{equation}
\end{small}

Certain types (e.g., positive, negative, and privative) of words should play more important roles than other words do in texts; therefore, their embeddings are also used in the attention layer. A new LSTM based on our lexicon embedding is proposed, as shown in Fig. 5. The attention layer and final dense vector of the network in Fig. 3(a) are calculated as follows.

\begin{equation}
\begin{array}{l}
{\alpha _t} = {\mathop{\rm softmax}\nolimits} ({W_\alpha }\left[ {\begin{array}{*{20}{c}}
{{h_t}}\\
{{l_t}}
\end{array}} \right] + {b_i})\\
\gamma  = \sum\limits_t {{\alpha _t}{h_t}}
\end{array}
\end{equation}
where $\alpha_t$ is the attention weight for the $t$-th input, lt is the lexicon embedding for key lexical words for the $t$-th input, and $\gamma$ is the final dense vector. Eq. (2) is used in the first-level LSTM.

\subsubsection{Embedding for POS}
POS is usually used as a key cue in sentiment analysis \cite{Cambria74}. To this end, we use additional lexicon embedding. The new lexicon embedding includes several major types of POS, namely, interrogative, exclamatory, and others. This new lexicon embedding is also applied to the attention layer. The motivation lies in that certain types of POS should play important roles in sentiment.

The proposed $\rho$-hot embedding is still applied to POS types in this study. According to our initial case studies, eight POS types are considered. They are noun, adjective, verb, pronoun, adverb, preposition, accessory, and others. The eight POS types are represented by the proposed $\rho$-hot encoding. We let $n$ in Eq. (6) be 10. The first three POS types are as follows.
\[\begin{array}{l}
{\rm{Noun:}}{[{{\bm{\rho }}_{1 \times 10}},{{\bf{0}}_{1 \times 10}},{{\bf{0}}_{1 \times 10}},{{\bf{0}}_{1 \times 10}},\cdots,{{\bf{0}}_{1 \times 10}}]^T}\\
{\rm{Adj}}:{[{{\bf{0}}_{1 \times 10}},{{\bm{\rho }}_{1 \times 10}},{{\bf{0}}_{1 \times 10}},{{\bf{0}}_{1 \times 10}},\cdots,{{\bf{0}}_{1 \times 10}}]^T}\\
{\rm{Verb}}:{[{{\bf{0}}_{1 \times 10}},{{\bf{0}}_{1 \times 10}},{{\bm{\rho }}_{1 \times 10}},{{\bf{0}}_{1 \times 10}},\cdots,{{\bf{0}}_{1 \times 10}}]^T}
\end{array}\]
When POS embedding is used, the attention layer and final outputs of the network in Eq. (3) become
\begin{equation}
\begin{array}{l}
{\alpha _t} = {\mathop{\rm softmax}\nolimits} ({W_\alpha }\left[ {\begin{array}{*{20}{c}}
{{h_t}}\\
\begin{array}{l}
{l_t}\\
{\eta _t}
\end{array}
\end{array}} \right] + {b_i})\\
\gamma  = \sum\limits_t {{\alpha _t}h_t}
\end{array}
\end{equation}
where $\eta_t$ is the lexicon embedding for key lexical words for the $t$-th input.

\subsubsection{Embedding for conjunction}

\begin{figure}[t]
\begin{center}
  \hspace{0in}\includegraphics[width= 0.438\textwidth, height = 146pt]{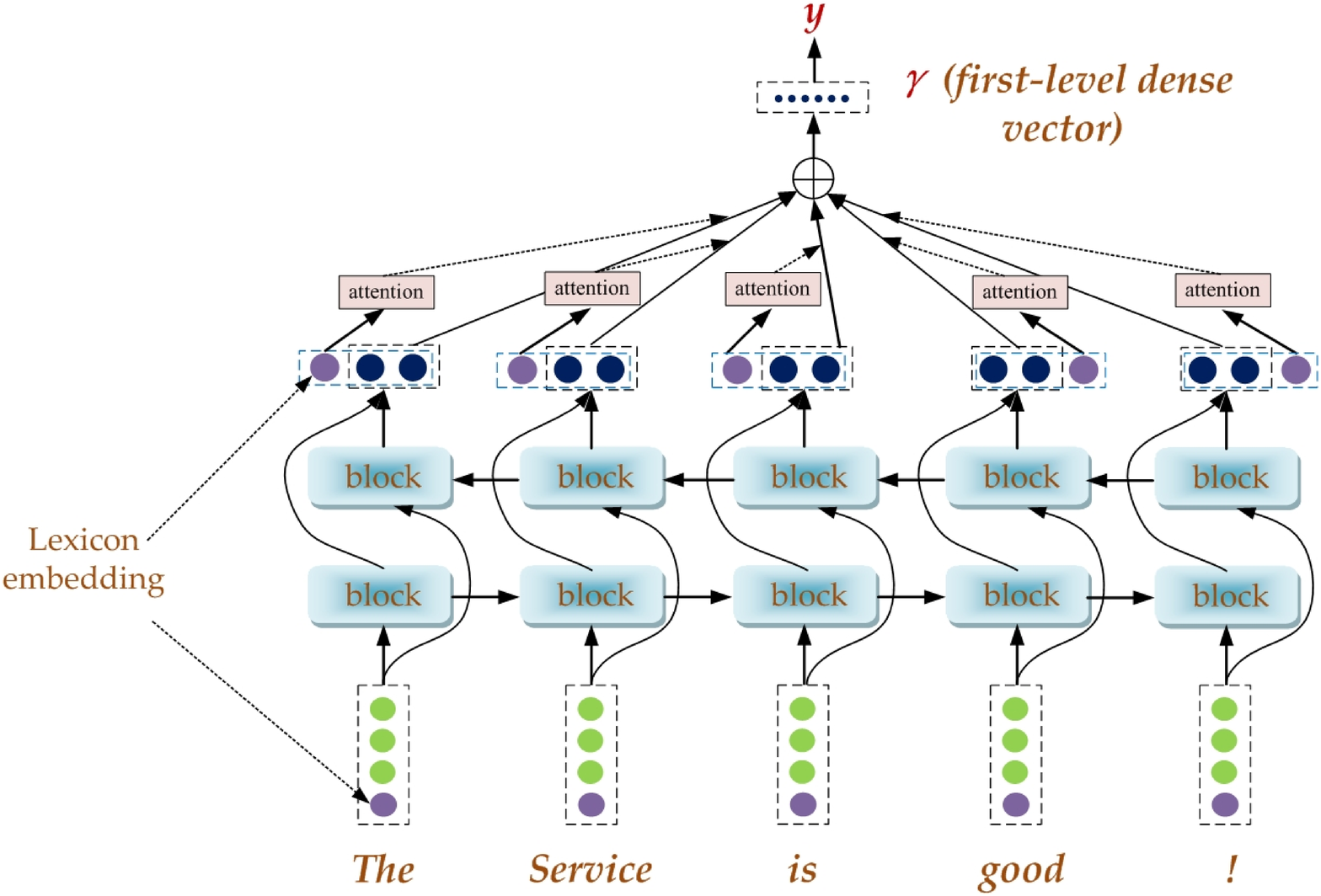}\\
  \vspace{-0.05in}\caption{The first-level LSTM with lexicon embedding in both the input and attention layers.}\vspace{-0.15in}
\end{center}
\end{figure}

Conjunction words play important roles in sentiment analysis \cite{Ding811}. For example, conjunctions such as ``but" and ``moreover" usually indicate the focus of texts and attract readers’ attention. Therefore, conjunctions are considered in the input of the second-level LSTM.

Once a set of conjunction words is compiled, $\rho$-hot embedding is used. In our experiments, the number of conjunction words is 169. Therefore, the parameter $n$ in Eq. (2) is set as 1.

When conjunction embedding is used for the second-level layer, the attention layer and final outputs of the network in Fig. 3(b) are calculated as follows.
\begin{equation}
\begin{array}{l}
{\beta _t} = {\mathop{\rm softmax}\nolimits} ({W_\beta }\left[ {\begin{array}{*{20}{c}}
{y_{_t}^{(1)}}\\
\begin{array}{l}
h{_t}^{(2)}\\
\omega _{_t}^{\rm{s}}\\
\omega _{_t}^e
\end{array}
\end{array}} \right] + {{b'}_i})\\
{\gamma ^{(2)}} = \sum\limits_t {{\beta _t}\left[ {\begin{array}{*{20}{c}}
{h_{_t}^{(2)}}\\
{y_{_t}^{(1)}}
\end{array}} \right]}
\end{array}
\end{equation}
where $\beta_t$ is the attention weight for the $t$-th input clause; $h_t^{(2)}$ is the hidden vector of the second-level LSTM; $\omega^s_t$ and $\omega^e_t$ are the conjunction embeddings for the first and last words in the $t$-th input clause, respectively; and $\gamma^{(2)}$ is the final dense vector used for the final classification.

\subsubsection{Differences between our and existing lexicon embedding}
Shin et al. \cite{Shin149} also embedded lexical information into sentiment analysis. Three major differences exist between our method and the method proposed by Shin et al. \cite{Shin149}.
\begin{itemize}
  \item The lexicon embedding proposed by Shin et al. us-es one-hot encoding, whereas the proposed method uses a new encoding strategy that can be considered a soft one-hot encoding.
  \item The lexicon embedding proposed by Shin et al. ex-tends the length of raw encoded vectors. However, the extension aims to keep the lengths of lexical and word embeddings equal. Their extension method also only relies on zero padding and is thus different from the proposed method.
  \item Only sentimental words are considered in the lexicon embedding proposed by Shin et al. On the contrary, sentimental words, POS, and conjunctions are considered in our work.
\end{itemize}

\begin{figure}[t]
\begin{center}
  \hspace{0in}\includegraphics[width= 0.496\textwidth, height = 180pt]{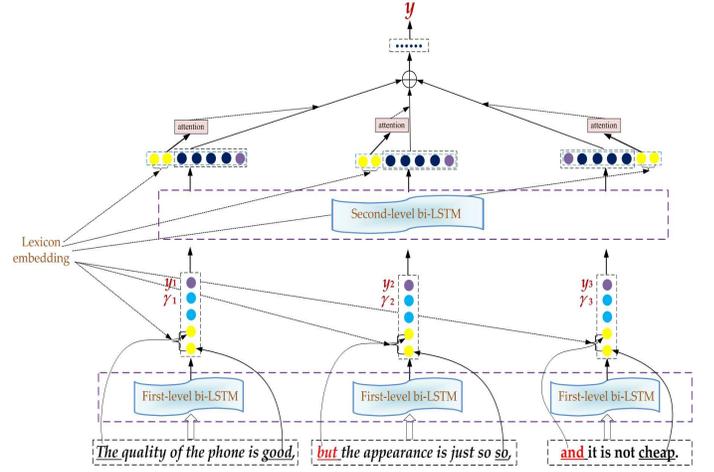}\\
  \vspace{-0.05in}\caption{The whole two-level LSTM network with lexicon embedding in both the input and attention layers.}\vspace{-0.15in}
\end{center}
\end{figure}

\subsection{The Learning Procedure}
The algorithmic steps of the entire learning procedure for the proposed $\rho$-hot lexicon embedding-based two-level LSTM (called $\rho$Tl-LSTM) are shown in \textbf{Algorithm} 1. In \textbf{Algorithm} 1, T1 refers to the training data that consist of clauses and the labels obtained in the first-stage labeling procedure. T2 refers to the training data that consist of sentences and the labels obtained in the second-stage labeling procedure. The structure of $\rho$Tl-LSTM is presented in Fig. 6.

\begin{algorithm}
\caption{$\rho$Tl-LSTM}
\begin{algorithmic}
\STATE \hspace{-0.16in} \textbf{Input}: Training sets T1 and T2; dictionary of key lexical words; POS for each word; dictionary of conjunction words; character/word embeddings for each character/word.\vspace{0.1in}
\STATE \hspace{-0.16in} \textbf{Output}: A trained two-level LSTM for sentiment classification.\vspace{0.1in}
\STATE \hspace{-0.16in} \textbf{Steps}: \vspace{0.0in}
\STATE \item[1.] Construct the embedding vector for each character (including punctuation) in the clauses in T1. The embeddings include the character/word and lexicon embeddings of each character/word;
\STATE \item[2.] Train the first-level LSTM on the basis of the input embedding vectors and labels of the T1 text clauses;
\STATE \item[3.] Run the learned first-level LSTM on each clause of the text samples in T2. Record the predicted score $y^{(1)}$ and the final dense vector $\gamma^{(1)}$ for each clause;
\STATE \item[4.] Construct the embedding vector for each clause in the text samples in T2. Each embedding vector consists of $y^{(1)}$, $\gamma^{(1)}$, and the lexicon embedding of conjunctions of each clause;
\STATE \item[5.] Train the second-level LSTM on the basis of the input embedding vectors and labels of the T2 text samples.\\
The first-level and second-level LSTM networks consist of the final two-level LSTM.
\end{algorithmic}
\end{algorithm}

The proposed two-level LSTM can be applied to texts with arbitrary languages. Word information is required in lexical construction regardless of whether character or word embedding is used. The reason is that the three types of lexicon embeddings are performed at the word level. Therefore, when character embedding is used, the lexicon embedding of each character is the lexicon embedding of the word containing it.


\section{EXPERIMENTS}
This section shows the evaluation of the proposed methodology in terms of the two-level LSTM network and each part of the lexicon embedding.
\subsection{Experimental Data and Labeling}
 We compile three Chinese text corpora from online data for three domains, namely, ``hotel", ``mobile phone (mobile)", and ``travel". All texts are about user reviews. Each text sample collected is first partitioned into clauses according to Chinese tokens\footnote{Token-based token is inaccurate for English text partition. Nevertheless, the segment results for Chinese texts are acceptable. A more reasonable way will be investigated in our future work.}. Three clause sets are subsequently obtained from the three text corpora.

The labels ``+1", ``0.5", and ``0" correspond to the three sentiment classes ``positive", ``neutral", and ``negative", respectively. The text data are labeled according to our two-stage labeling strategy.
\begin{itemize}
  \item In the first stage, only one user is invited to label each clause sample as the sentiment orientations for clauses (or sub-sentences) are easy to label.
  \item In the second stage, five users\footnote{Five graduate students, including three males and two females, were invited to label the data.} are invited to label each text sample in the three raw data sets. The average score of the five users on each sample is calculated. Samples with average scores located in [0.6, 1] are labeled as ``positive". Samples with average scores located in [0, 0.4] are labeled as ``negative". Others are labeled as ``neutral". The details of the labeling results are shown in Table 1.
\end{itemize}

\begin{table}[h]
\renewcommand\arraystretch{1.18}
\begin{center}
\caption{\normalsize Details of the three data corpora. Each corpus consists of raw samples (sentences or paragraphs) and partitioned clauses (sub-sentences).}
{\footnotesize \begin{tabular}{|c|c |c|c|}\hline
\multicolumn{2}{|c|}{Data corpus}  &  raw  & clauses\\\hline\hline
\multirow{4}{*}{Travel}&Pos.&1567&3490\\
&Neu.&576&5168\\
&Neg.&1957&2633\\
&Total&4100&11291\\\hline
\multirow{4}{*}{Hotel}&Pos.&1586&3987\\
&Neu.&401&2123\\
&Neg.&1838&5154\\
&Total&3825&11264\\\hline
\multirow{4}{*}{Mobile}&Pos.&1400&2788\\
&Neu.&589&2375\\
&Neg.&1494&2955\\
&Total&3483&8118\\\hline
\end{tabular}}
\end{center}
\end{table}

All the training and test data and the labels are available online\footnote{https://github.com/Tju-AI/two-stage-labeling-for-the-sentiment-orientations/tree/master/data}.

In our experiments, the five types of key lexical words introduced in Subsection 3.3.2 are manually constructed. The details of the five types of words are listed in Table 2\footnote{The five types of key lexical words are also available and introduced in our Github project page.}. The conjunction words are also manually constructed. The number of conjunction words used in the experiments is 169.

\begin{table}[h]
\renewcommand\arraystretch{1.18}
\begin{center}
\caption{\normalsize Numbers of five types of key lexical words.}
{\footnotesize \begin{tabular}{|c||c ||c|   |c|}\hline
Data corpus&Travel&Hotel&Mobile\\\hline\hline
Positive&366&254&358\\
Negative&327&194&382\\
Privative&61&61&61\\
Interrogative&48&48&48\\
Suppositive&18&18&18\\\hline
\end{tabular}}
\end{center}
\end{table}

In each experimental run, the training set is compiled on the basis of the training data listed in Table 1. The compiling rule is specified before each experimental run. The test data are fixed to facilitate experimental duplication and comparison by other researchers.

\subsection{Competing Methods and Parameter Setting}

In our experiments, three competing algorithms, namely, BOW, CNN, and (conventional) LSTM, are used.

For BOW, term frequency-inverse document frequency is utilized to construct features. Ridge regression \cite{Dobriban} is used as a classifier. For CNN, a three-channel CNN is used. For LSTM, one-layer and two-layer bi-LSTM with attention are adopted, and the results of the network with superior performance are presented. CNN and LSTM are performed on TensorFlow, and default parameter settings are followed.

The key parameters are searched as follows. The embedding dimensions of characters and words are searched in [100, 150, 200, 250, 300]. The parameter $n$ in $\rho$-hot encoding is searched in $[1, 3, \cdots, 15]$.

\subsection{Results}

\subsubsection{Results of existing baseline methods}
In this subsubsection, each of the three raw data sets (associated with their labels) shown in Table 1 is used. The clause data are not used. In other words, the training data used in this subsubsection are the same as those used in previous studies. For each data corpus, 1000 raw data samples are used as the test data, and the rest are used as the training data.
The involved algorithms are detailed as follows.
\begin{itemize}
  \item \textbf{CNN-C} denotes the CNN with (Chinese) character embedding.
  \item \textbf{CNN-W} denotes the CNN with (Chinese) word embedding.
  \item \textbf{CNN-Lex-C} denotes the algorithm which also integrates polar words in CNN which is proposed by Shin et al.\cite{Shin149}. The (Chinese) character embedding is used.
  \item \textbf{CNN-Lex-W} denotes the algorithm which also integrates polar words in CNN which is proposed by Shin et al.\cite{Shin149}. The (Chinese) word embedding is used.
  \item \textbf{Bi-LSTM-C} denotes the BI-LSTM with (Chinese) character embedding.
  \item \textbf{Bi-LSTM-W} denotes the Bi-LSTM with (Chinese) word embedding.
  \item \textbf{Lex-rule} denotes the rule-based approach shows in Fig. 1. This approach is unsupervised.
  \item \textbf{BOW} denotes the conventional algorithm which is based of bag-of-words features.
\end{itemize}

The accuracies of the above algorithms are listed in Table 3. Overall, Bi-LSTM outperforms CNN and BOW. This conclusion is in accordance with the conclusion that RNN performs efficiently against CNN in a broad range of natural language processing (NLP) tasks on the basis of extensive comparative studies \cite{Yin1702}. In addition, CNN-lex outperforms CNN under both character and word embeddings, which suggests that lexicon cues are useful in sentiment analysis. Lex-rule achieves the lowest accuracies on all the three data sets. Considering that the performances of (traditional) CNN, Lex-rule, and BOW are relatively poor, they are not applied in the remaining parts.

\begin{table}[h]
\renewcommand\arraystretch{1.28}
\begin{center}
\caption{\normalsize The classification accuracies of existing algorithms on raw samples.}
{\footnotesize \begin{tabular}{|c||c ||c||c|}\hline
Data corpus&Travel&Hotel&Mobile\\\hline\hline
CNN-C&0.723&	0.698&	0.727\\
CNN-W&0.731&	0.729&	0.748\\
CNN-Lex-C&0.744&0.734&0.731\\
CNN-Lex-W&0.758&0.764&0.755\\
Bi-LSTM-C&\textbf{0.754}&	0.753&	0.805\\
Bi-LSTM-W&0.746&	\textbf{0.785}&	\textbf{0.809}\\
Lex-rule&0.556&0.539&0.684\\
BOW&0.713&	0.678&	0.702\\\hline
\end{tabular}}
\end{center}
\end{table}

\subsubsection{Results of two-level LSTM without lexicon embedding}

In this experimental comparison, the proposed two-level LSTM is evaluated, whereas lexicon embedding is not used in the entire network. The primary goal is to test whether the introduced two-stage labeling and the two-level network structure are useful for sentiment analysis.

The raw and clause data listed in Table 1 are used to perform the two-level LSTM. Tl-LSTM denotes the two-level LSTM. ``R+C" refer to the mixed data of raw and clause data. The test data are still the 1000 samples used in section 4.3.1 for each corpus. Table 4 shows the classification accuracies. To ensure that the results differ from those in Table 3, we explicitly add ``R+C" after each algorithm in Table 4. In the last line of Table 4, the base results for each corpus in Table 3 are also listed.

\begin{table}[h]
\renewcommand\arraystretch{1.28}
\begin{center}
\caption{\normalsize The classification accuracies of two-level LSTM without lexicon embedding.}
{\footnotesize \begin{tabular}{|c||c ||c||c|}\hline
Data corpus&Travel&Hotel&Mobile\\\hline\hline
Tl-LSTM-C(R+C)&\textbf{0.8010}&	\textbf{0.8130}&	\textbf{0.8200}\\
Tl-LSTM-W(R+C)&0.7700&	0.7720&	0.8200\\
CNN-Lex-C (R+C)&0.7550&0.7700&0.7490\\
CNN-Lex-W (R+C)&0.7730&0.7760&0.7860\\
Bi-LSTM-C (R+C)&0.7810&	0.7840&	0.8170\\
Bi-LSTM-W (R+C)&0.7620&	0.7890&	0.8130\\
Baseline (best in Table 3)&0.7540&	0.7850&	0.8090\\\hline
\end{tabular}}
\end{center}
\end{table}

On all the three data corpora, the proposed two-level network (without lexicon embedding) with character embedding, Tl-LSTM-C, outperforms all the other involved algorithms. On the travel and the mobile corpora, TI-LSTM-W outperforms Bi-LSTM-W. The results in Table 4 indicate that the performances of Tl-LSTM on the mixed training and test data (R+C) are better than those of Bi-LSTM. This comparison indicates that the proposed two-level LSTM is effective.

In addition, for the involved algorithms, most results  achieved on ``R+C" are better than the best results only achieved on `R'listed in Table 3. This comparison suggests that the introduced two-stage labeling is useful.

The results also show that in the two-level LSTM, character embedding is more effective than word embedding.

\subsubsection{Results of lexicon embedding-based two-level LSTM}

In this experimental run, lexicon embedding is used in the proposed two-level LSTM or $\rho$Tl-LSTM. Table 5 shows the results. The optimal parameter $n$ is about 11.

The performances of TI-LSTM with lexicon embedding (i.e., $\rho$Tl-LSTM) are consistently better than those of TI-LSTM without lexicon embedding (i.e., Tl-LSTM) listed in Table 5. The improved accuracies of $\rho$TI-LSTM over Tl-LSTM on the three data corpora are explicitly listed in Table 6.
\begin{table}[h]
\renewcommand\arraystretch{1.28}
\begin{center}
\caption{\normalsize The classification accuracies of two-level LSTM with lexicon embedding.}
{\footnotesize \begin{tabular}{|c||c ||c||c|}\hline
Data corpus&Travel&Hotel&Mobile\\\hline\hline
$\rho$Tl-LSTM-C (R+C)&\textbf{0.816}&\textbf{0.837}&0.826\\
$\rho$Tl-LSTM-W (R+C)&0.800&0.810&\textbf{0.841}\\\hline
\end{tabular}}
\end{center}
\end{table}

\newcommand{\tabincell}[2]{\begin{tabular}{@{}#1@{}}#2\end{tabular}}
\begin{table}[h]
\renewcommand\arraystretch{1.28}
\begin{center}
\caption{\normalsize The accuracy improvement of two-level LSTM when lexicon embedding was used over those of two-level LSTM without lexicon embedding. The values of the last row are the accuracy improvement over the highest accuracies on each data corpus with existing algorithms.}
{\footnotesize \begin{tabular}{|c ||c ||c |}\hline
Travel&Hotel&Mobile\\\hline\hline
+0.6\%&+2.4\%&+0.6\%\\\hline
+3\%&+2.8\%&+2.1\%\\\hline
\tabincell{c}{+6.02\% \\(Compared with\\ the highest in Tbl. 3)}&\tabincell{c}{+5.2\%\\(Compared with \\the highest in Tbl. 3)} &\tabincell{c}{+5.01\%\\(Compared with\\ the highest in Tbl. 3)}\\\hline
\end{tabular}}
\end{center}
\end{table}

\subsection{Disussion}
The experimental evaluation discussed in Subsection 4.3 verifies the effectiveness of the proposed method, $\rho$Tl-LSTM. Unlike the conventional RNN, $\rho$Tl-LSTM contains lexicon embedding that consists of new technique and components, including $\rho$-hot encoding, embedding for polar words, embedding for POS, and embedding for conjunctions. Therefore, this subsection evaluates the performances of the involved technique and embeddings separately.

\subsubsection{The effect of different parameters on $\rho$-hot encoding}
Our $\rho$-hot encoding differs from one-hot encoding in two aspects. The first aspect is that the nonzero values in one-hot encoding are only equal to 1, whereas the nonzero values in $\rho$-hot encoding are $\rho$. The second aspect is that only one element in one-hot encoding is nonzero, whereas n elements in $\rho$-hot encoding are nonzero.

In this experiment, we test whether $\rho$-hot encoding is useful in two experimental runs. In the first run, the value of $\rho$ is manually set to 0.5 and 1 in the experimental run without optimization. The parameter $n$ in Eq. (6) is set as 15. The classification accuracies vary according to different $\rho$ values on all the three data corpora. When $\rho$ equals 1, the accuracies are the lowest in most cases shown in Fig. 7.

The results shown in Fig. 7 indicate that the value of $\rho$ does affect the performance of the entire network. Consequently, the classical one-hot encoding, which fixes the value of nonzero elements as 1, is ineffective. In our experiments, the learned value of $\rho$ is approximate 0.4.

In the second run, the performances under different $n$ (i.e., 1, 5, 10, 15) are tested. Table 7 shows the comparison results. The value of $n$ does affect the performance of the entire network, thereby indicating that the introduction of the $n$-duplicated strategy in encoding is effective. In the experiments, when $n$ is increasing, the accuracies first increase and then decrease. The main reason may lie in the fact that when $n$ becomes large, the proportion of lexicon embedding becomes large accordingly. An over-length input feature vector may incur ``curse of dimensionality" and thus weaken the performance of the proposed two-level network.

\begin{figure}[h]
\centerline{\includegraphics[width=0.3\textwidth]{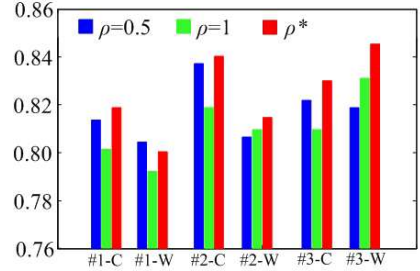}}
 \vspace{-0.1in}\caption{Classification accuracies under different $\rho$ values. \#1-C and \#1-W represent $\rho$Tl-LSTM-C and $\rho$Tl-LSTM-W on the first (travel) corpus, respectively; \#2-C and \#2-W represent $\rho$Tl-LSTM-C and $\rho$Tl-LSTM-W on the second (hotel) corpus, respectively; \#3-C and \#3-W represent $\rho$Tl-LSTM-C and $\rho$Tl-LSTM-W on the third (hotel) corpus, respectively.}
\label{f7}
\end{figure}

\begin{table}[h]
\renewcommand\arraystretch{1.28}
\begin{center}
\caption{\normalsize The accuracies of $\rho$Tl-LSTM with different $n$ values in $\rho$-hot encoding.}
{\footnotesize \begin{tabular}{|c||c||c||c||c|}\hline
&$n$&Travel&Hotel&Mobile\\\hline
\multirow{4}{*}{$\rho$Tl-LSTM-C}&1&0.790&	0.814	&0.807\\
&5&0.799&0.824&0.825\\
&10&0.803&0.831&0.814\\
&searched&0.816&0.837&0.826\\\hline
\multirow{4}{*}{$\rho$Tl-LSTM-W}&1&0.792&0.798&0.820\\
&5&0.804&0.810&0.827\\
&10&0.800&0.802&0.836\\
&searched&0.800&0.810&0.841\\\hline
\end{tabular}}
\end{center}
\end{table}

\subsubsection{The effect of polar words}

In this experimental run, we test whether the labeled polar (negative and positive) words do affect the performance of the entire method when they are used in lexicon embedding. To this end, we order the polar words according to their frequencies in the training data. Top 0\%, 50\%, 100\% polar words are used. The corresponding classification accuracies are depicted in Fig. 8.

In most cases, the accuracies are the lowest when no polar words are used in the lexicon embedding. When all polar words are used, the proposed network achieves the highest accuracies.

In the experiment, only one user is invited to manually compile the dictionary for a data corpus. One and a half hour is needed for each data corpus. In our viewpoint, it is worth manually compiling the polar words for sentiment analysis by considering the performance improvement and time-consumption.

\begin{figure}[t]
\centerline{\includegraphics[width=0.30\textwidth]{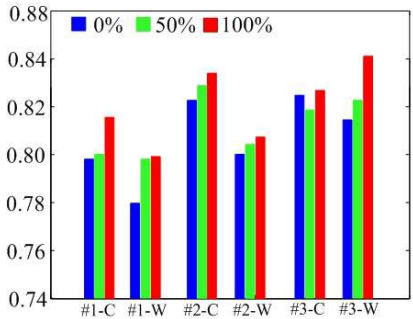}}
\vspace{-0.1in}\caption{Classification accuracies under different proportions of polar words. \#1-C and \#1-W represent $\rho$Tl-LSTM-C and $\rho$Tl-LSTM-W on the first (travel) corpus, respectively; \#2-C and \#2-W represent $\rho$Tl-LSTM-C and $\rho$Tl-LSTM-W on the second (hotel) corpus, respectively; \#3-C and \#3-W represent $\rho$Tl-LSTM-C and $\rho$Tl-LSTM-W on the third (hotel) corpus, respectively.}
\label{f9}
\end{figure}

\subsubsection{The effect of POS cues}
In this experimental run, we test whether POS cues do play positive roles in the entire model. To this end, we remove POS in the lexicon embedding of the proposed method. The results are shown in Fig. 9.

 In most cases, the accuracies with POS embedding are greater than those without POS embedding, thereby indicating that the application of POS to lexicon embedding is useful.

\begin{figure}[t]
\begin{center}
  \hspace{0in}\includegraphics[width= 0.3\textwidth]{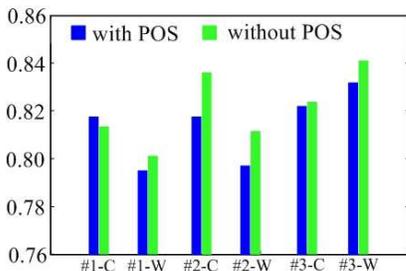}\\
  \vspace{-0.205in}\caption{Classification accuracies with and without POS in lexicon embedding. \#1-C and \#1-W represent $\rho$Tl-LSTM-C and $\rho$Tl-LSTM-W on the first (travel) corpus, respectively; \#2-C and \#2-W represent $\rho$Tl-LSTM-C and $\rho$Tl-LSTM-W on the second (hotel) corpus, respectively; \#3-C and \#3-W represent $\rho$Tl-LSTM-C and $\rho$Tl-LSTM-W on the third (hotel) corpus, respectively.}\vspace{-0.15in}
\end{center}
\end{figure}

\subsubsection{The effect of conjunction cues}
In this experimental run, we test whether conjunction cues do play positive roles in the entire model. To this end, the lexicon embedding for conjunction words is also removed from the proposed method. The results are shown in Fig. 10.

The algorithm with conjunction embedding outperforms that without conjunction embedding consistently, thereby indicating that the application of conjunction to lexicon embedding is useful.

\begin{figure}[t]
\begin{center}
  \hspace{0in}\includegraphics[width= 0.3\textwidth]{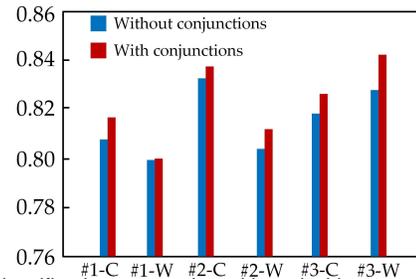}\\
  \vspace{-0.205in}\caption{Classification accuracies with and without conjunction in lexicon embedding. \#1-C and \#1-W represent $\rho$Tl-LSTM-C and $\rho$Tl-LSTM-W on the first (travel) corpus, respectively; \#2-C and \#2-W represent $\rho$Tl-LSTM-C and $\rho$Tl-LSTM-W on the second (hotel) corpus, respectively; \#3-C and \#3-W represent $\rho$Tl-LSTM-C and $\rho$Tl-LSTM-W on the third (hotel) corpus, respectively.}\vspace{-0.15in}
\end{center}
\end{figure}

\section{Conclusion}
High-quality labels are crucial for learning systems. Nevertheless, texts with mixed sentiments are difficult for humans to label in text sentiment classification. In this study, a new labeling strategy is introduced to partition texts into those with pure and mixed sentiment orientations. These two categories of texts are labeled using different processes. A two-level network is accordingly proposed to utilize the two labeled data in our two-stage labeling strategy. Lexical cues (e.g., polar words, POS, conjunction words) are particularly useful in sentiment analysis. These lexical cues are used in our two-level network, and a new encoding strategy, that is, $\rho$-hot encoding, is introduced. $\rho$-hot encoding is motivated by one-hot encoding. However, the former alleviates the drawbacks of the latter. Three Chinese sentiment text data corpora are compiled to verify the effectiveness of the proposed methodology. Our proposed method achieves the highest accuracies on these three data corpora.

The proposed two-level network and lexicon embedding can also be applied to other types of deep neural networks. In our future work, we will extend our main idea into several networks and text mining applications.


%

\appendices

\ifCLASSOPTIONcompsoc
  \section*{Acknowledgments}
\else
  \section*{Acknowledgment}
\fi
The authors wish to thank Zefeng Han, Qing Yin, Lei Yang, Xiaonan Wang, Nan Chen, Rujing Yao, Lihong Guo, Pinglong Zhao for the labeling of the experimental data.

\ifCLASSOPTIONcaptionsoff
  \newpage
\fi


\begin{thebibliography}{99}

\bibitem{LiuBook2015}
B.~Liu, \emph{Sentiment Analysis: Mining Opinions, Sentiments, and
  Emotions}.\hskip 1em plus 0.5em minus 0.4em\relax Cambridge University Press,
  2015.

\bibitem{Alexander17}
A.~Pak and P.~Paroubek, ``Twitter as a corpus for sen-timent analysis and
  opinion mining,'' in \emph{Proceedings of International Conference on
  Language Resources and Evaluation (LREC)}, 2010, pp. 17--23.

\bibitem{Kim1746}
Y.~Kim, ``Convolutional neural networks for sentence classification,'' in
  \emph{Proceedings of International Conference on Empirical Methods in Natural
  Language (EMNLP)}, 2014, pp. 1746--1751.

\bibitem{Taboada267}
M.~Taboada, J.~Brooke, M.~Tofiloski, K.~Voll, and M.~Stede, ``Lexicon-based
  methods for sentiment analysis,'' \emph{Computational Linguistics}, vol.~37,
  no.~2, pp. 267--307, 2011.

\bibitem{Hochreiter1735}
S.~Hochreiter and J.~Schmidhuber, ``Long short-term memory,'' \emph{Neural
  Computation}, vol.~9, no.~8, pp. 1735--1780, 1997.

\bibitem{ZhaoGCHCWW18}
W.~Zhao, Z.~Guan, L.~Chen, X.~He, D.~Cai, B.~Wang, and Q.~Wang,
  ``Weakly-supervised deep embedding for product review sentiment analysis.''
  \emph{IEEE Trans. Knowl. Data Eng.}, vol.~30, no.~1, pp. 185--197, 2018.

\bibitem{Hu168}
M.~Hu and B.~Liu, ``Mining and summarizing customer reviews,'' in
  \emph{Proceedings of ACM SIGKDD International Conference Knowledge Discovery
  Data Mining (KDD)}, 2004, pp. 168--177.

\bibitem{Mullen412}
T.~Mullen and N.~Collier, ``Sentiment analysis using support vector machines
  with diverse information sources,'' in \emph{Proceedings of International
  Conference on Empirical Methods in Natural Language (EMNLP)}, 2004, pp.
  412--418.

\bibitem{Paltoglou1386}
G.~Paltoglou and M.~Thelwall, ``A study of information retrieval weighting
  schemes for sentiment analysis,'' in \emph{Proceedings of Annual Meeting of
  the Association for Computational Linguistics (ACL)}, 2010, pp. 1386--1395.

\bibitem{Glorot513}
X.~Glorot, A.~Bordes, and Y.~Bengio, ``Domain adaptation for large-scale
  sentiment classification: A deep learning approach,'' in \emph{Proceedings of
  International Conference on Machine Learning (ICML)}, 2011, pp. 513--520.

\bibitem{Tang3298}
D.~Tang, B.~Qin, X.~Feng, and T.~Liu, ``Effective lstms for target dependent
  sentiment classification,'' in \emph{Proceedings of International Conference
  on Computational Linguistics (COLING)}, 2016, pp. 3298--3307.

\bibitem{Baccianella2200}
A.~E. Baccianella, Stefano, and F.~Sebastiani, ``Senti-wordnet 3.0: An enhanced
  lexical resource for sentiment analy-sis and opinion mining,'' in
  \emph{Proceedings of International Conference on Language Resources and
  Evaluation (LREC)}, 2010, pp. 2200--2204.

\bibitem{Lai5}
S.~Lai, K.~Liu, L.~Xu, and J.~Zhao, ``How to generate a good word embedding,''
  \emph{IEEE Intelligent Systems}, vol.~31, no.~6, pp. 5--14, 2015.

\bibitem{Cho1724}
K.~Cho, B.~van Merrienboer, C.~Gulcehre, D.~Bahdanau, F.~Bougares, H.~Schwenk,
  and Y.~Bengio, ``Learning phrase representations using rnn encoder-decoder
  for statistical machine translation,'' in \emph{Proceedings of International
  Conference on Empirical Methods in Natural Language (EMNLP)}, 2014, pp.
  1724--1734.

\bibitem{Yang645}
M.-C. Yang, N.~Duan, M.~Zhou, and H.-C. Rim, ``Joint relational embeddings for
  knowledge-based question answering,'' in \emph{Proceedings of International
  Conference on Empirical Methods in Natural Language (EMNLP)}, 2014, pp.
  645--650.

\bibitem{Dai1507}
A.~M. Dai, C.~Olah, and Q.~V. Le, ``Document embedding with paragraph
  vectors,'' \emph{CoRR}, vol. abs/1507.07998, 2015.

\bibitem{Tang1422}
D.~Tang, B.~Qin, and T.~Liu, ``Document modeling with gat-ed recurrent neural
  network for sentiment classification,'' in \emph{Proceedings of International
  Conference on Empirical Methods in Natural Language (EMNLP)}, 2015, pp.
  1422--1432.

\bibitem{Kalchbrenner655}
N.~Kalchbrenner, E.~Grefenstette, and P.~Blunsom, ``A convolutional neural
  network for modelling sentences,'' in \emph{Proceedings of Annual Meeting of
  the Association for Computational Linguistics (ACL)}, 2014, pp. 655--665.

\bibitem{Socher129}
R.~Socher, C.~C.-Y. Lin, A.~Y. Ng, and C.~D. Manning, ``Parsing natural scenes
  and natural lan-guage with recursive neural networks,'' in \emph{Proceedings
  of International Conference on Machine Learning (ICML)}, 2011, pp. 129--136.

\bibitem{Zhang3087}
M.~Zhang, Y.~Zhang, and D.~Vo, ``Gated neural networks for targeted sentiment
  analysis,'' in \emph{Proceedings of AAAI Conference (AAAI)}, 2016, pp.
  3087--3093.

\bibitem{Santos69}
C.~Santos, N.~Dos, and M.~Gattit, ``Deep convolu-tional neural networks for
  sentiment analysis of short texts,'' in \emph{Proceedings of International
  Conference on Computational Linguistics (COLING)}, 2014, pp. 69--78.

\bibitem{Gers2451}
F.~A. Gers, J.~Schmidhuber, and F.~Cummins, ``Learning to forget: continual
  prediction with lstm,'' \emph{Neural Computation}, vol.~12, no.~10, pp.
  2451--2471, 2000.

\bibitem{Zhou247}
X.~Zhou, X.~Wan, and J.~Xiao, ``Attention-based lstm network for cross-lingual
  sentiment classification,'' in \emph{Proceedings of International Conference
  on Empirical Methods in Natural Language (EMNLP)}, 2016, pp. 247--256.

\bibitem{Qian1611}
Q.~Qian, M.~Huang, J.~Lei, and X.~Zhu, ``Linguistically regularized lstms for
  sentiment classification,'' \emph{Proceedings of Annual Meeting of the
  Association for Computational Linguistics (ACL)}, vol. 1679--1689, 2017.

\bibitem{Shin149}
B.~Shin, T.~Lee, and J.~D. Choi, ``Lexicon integrated cnn models with attention
  for sentiment analysis,'' in \emph{Proceedings of WAS-SA@EMNLP}, 2017, pp.
  149--158.

\bibitem{Schouten813}
K.~Schouten and F.~Frasincar, ``Survey on aspect-level sentiment analysis,''
  \emph{IEEE Transactions on Knowledge and Data Engineering}, vol.~28, no.~3,
  pp. 813--830, 2016.

\bibitem{Narayanan180}
R.~Narayanan, B.~Liu, and A.~Choudhary, ``Sentiment analysis of conditional
  sentences,'' in \emph{Proceedings of International Conference on Empirical
  Methods in Natural Language (EMNLP)}, 2009, pp. 180--189.

\bibitem{Cambria74}
E.~Cambria, S.~Poria, A.~F. Gelbukh, and M.~Thelwall, ``Sentiment analysis is a
  big suitcase,'' \emph{IEEE Intelligent Systems}, vol.~32, no.~6, pp. 74--80,
  2017.

\bibitem{Ding811}
X.~Ding and B.~Liu, ``The utility of linguistic rules in opinion mining,'' in
  \emph{Proceedings of the 30th Annual international ACM SIGIR Conference
  (SIGIR)}, 2007, pp. 811--812.

\bibitem{Dobriban}
E.~Dobriban and S.~Wager, ``High-dimensional asymptotics of prediction: Ridge
  regression and classification,'' \emph{The Annals of Statistics}, vol.~46,
  no.~1, pp. 247--279, 2018.

\bibitem{Yin1702}
W.~Yin, K.~Kann, M.~Yu, and H.~Sch{\"{u}}tze, ``Comparative study of {CNN} and
  {RNN} for natural language processing,'' \emph{CoRR}, vol. abs/1702.01923,
  2017.
  
\end{thebibliography}
\end{document}